\PassOptionsToPackage{table}{xcolor}
\documentclass{article} % For LaTeX2e
\usepackage{iclr2025_conference,times}

\usepackage{hyperref}
\usepackage{xcolor} 

\usepackage{url}
\usepackage{booktabs} 
\usepackage{multirow}
\usepackage{graphicx}
\usepackage{algorithm}
\usepackage{algpseudocode}
\usepackage{amsmath}
\usepackage{enumerate}
\usepackage{amsmath,amsfonts,bm}

\definecolor{lightblue}{RGB}{89, 149, 237}
\hypersetup{
    colorlinks=true,     % Enable colored links
    citecolor=lightblue,      % Color of citations
}

\definecolor{lightgray}{RGB}{240,240,240}
\definecolor{beige}{RGB}{242,226,210}

\title{Anchored Alignment for Self-Explanations Enhancement}

\author{\textbf{Luis Felipe Villa-Arenas}\textsuperscript{1,2}\thanks{Code repository: \url{https://github.com/felipevillaarenas/anchored-alignment}}, 
    \textbf{Ata Nizamoglu}\textsuperscript{1}, 
    \textbf{Qianli Wang}\textsuperscript{1},\\ 
    \textbf{Sebastian Möller}\textsuperscript{1} 
    \& 
    \textbf{Vera Schmitt}\textsuperscript{1}\\
    \textsuperscript{1}Department of Computer Science, Technische Universität Berlin, Germany\\
    \textsuperscript{2}Deutsche Telekom AG, Germany\\
    \texttt{\{luis-felipe.villa-arenas\}@campus.tu-berlin.de}
}

\iclrfinalcopy % Uncomment for camera-ready version, but NOT for submission.
\begin{document}

\maketitle
\begin{abstract}

In this work, we introduce a methodology for alignment designed to enhance the ability of large language models (LLMs) to articulate their reasoning—\textit{self-explanation}—even in the absence of annotated rationale explanations. Our alignment methodology comprises three key components: explanation quality assessment, self-instruction dataset generation, and model alignment. Additionally, we present a novel technique called \textit{Alignment with Anchor Preference Pairs}, which improves the selection of preference pairs by categorizing model outputs into three groups: consistently correct, consistently incorrect, and variable. By applying tailored strategies to each category, we enhance the effectiveness of Direct Preference Optimization (DPO). Our experimental results demonstrate that this approach significantly improves explanation quality while maintaining accuracy compared to other fine-tuning strategies.

\end{abstract}

\section{Introduction}

Large language models (LLMs) have demonstrated remarkable capabilities across various tasks. However, fine-tuning these models for specific applications often leads to a critical trade-off: improvements in one area may compromise the model's generalization capabilities \citep{yang_unveiling_2024, kirk_understanding_2024}. In our study, we aim to enhance a secondary task—specifically, the model's ability to articulate reasoning processes in natural language, a skill known as \textit{self-explanation} \citep{madsen_are_2024}—in parallel with the primary task, despite the constraint of not having human-annotated rationales.

The lack of annotated data of both high- and low-quality explanations can be framed in the context of model aligning without human preference data. Recent research has explored ways to align LLMs without direct human input. Some approaches generate self-instruct data to fine-tune models \citep{wang_self-instruct_2023, chen_alpagasus_2023, gulcehre_reinforced_2023}, while others, like \cite{bai_constitutional_2022, yuan_self-rewarding_2024, wu_meta-rewarding_2024}, use LLM-generated feedback to train reward models.  Building on these advancements, we propose an end-to-end approach to align LLMs on classification tasks while also ensuring the generation of high-quality self-explanations, even without annotated data for this secondary task. Our approach integrates three core components: evaluating generated explanations, creating self-instruct datasets, and aligning the model. Additionally, we introduce \textit{Alignment with Anchor Preference Pairs}, a method that improves preference pair selection by categorizing model responses into three groups: consistently correct, consistently incorrect, and variable. For each category, we apply tailored strategies to construct preference pairs, which are then used in the Direct Preference Optimization (DPO) phase \citep{rafailov_direct_2023}. Our results demonstrate that this method consistently improves explanation quality, mitigating the degradation caused by SFT. Moreover, we show that using anchor preference pairs outperforms self-alignment strategies that rely solely on judge-based evaluations for preference pair selection.

Our contributions are summarized as follows:

\begin{enumerate}[\itshape(i).] \item We introduce a framework for the qualitative assessment of self-explanations, designed to evaluate how effectively the model conveys its reasoning.

\item We analyze how  supervised fine-tuning for classification tasks affects the quality of self-explanations. Our findings demonstrate that while SFT improves classification accuracy, it often reduces explanation quality, underscoring the need for improved alignment strategies.

\item We propose a novel method, \textit{Alignment with Anchor Preference Pairs}, for constructing high-quality preference pairs when building self-instruct datasets. This method uses the model's behavior on each input prompt to apply specific strategies while creating preference pairs. Our approach consistently outperforms other methods that rely solely on judge-based evaluations for selecting preference pairs.

\item We develop an end-to-end methodology for aligning LLMs to downstream classification tasks while maintaining the quality of their self-explanations, even in the absence of explanation-rich datasets.

 \end{enumerate}

\section{A Framework for Qualitative Assessment of Self-Explanations} \label{sec: Self-Explanations Evaluation}

\subsection{Quality Criteria for Effective Self-Explanations} 
\label{sec: Quality Criteria for Effective Self-Explanations}

To assess self-explanation quality, we focus on the model's ability to effectively communicate its reasoning. This approach differs from previous work that emphasized trustworthiness metrics such as faithfulness \citep{madsen_faithfulness_2024, madsen_are_2024, lanham_measuring_2023, lyu_faithful_2023, turpin_language_2023, parcalabescu_measuring_2024} and truthfulness \citep{zhang_self-alignment_2024, sharma_towards_2023, burns_discovering_2022, joshi_personas_2024}. We evaluate self-explanations based on the following criteria:

\begin{enumerate}
    \item \textbf{Logical coherence}: The explanation should follow a clear and logical reasoning process, with all components cohesively connected to form a unified, non-contradictory narrative.
    \item \textbf{Clarity}: The explanation must present ideas clearly and precisely, using appropriate terminology to effectively communicate complex concepts without unnecessary complexity.
    \item \textbf{Relevance}: The explanation should comprehensively address the task at hand, directly answering the specific context or requirements without omitting critical information.
    \item \textbf{Depth of argumentation}: The explanation must provide strong reasoning and credible evidence to support its conclusions, reflecting a deep understanding of the task.
    \item \textbf{Factual accuracy}: This criterion assesses the correctness of individual claims within the explanation. While related to truthfulness, factual accuracy focuses on whether specific statements align with established knowledge.
\end{enumerate}

\subsection{Self-Explanations Evaluation Methodology}
\label{sec:Self-Explanations Evaluation Methodology}

Let $\mathcal{M}$ represent a large language model tasked with generating responses for a classification problem. Each response consists of two components: a self-explanation, denoted as $\varepsilon_i$, and a predicted classification label, $\hat{y}_i$, corresponding to an input prompt $x_i$. The self-explanation $\varepsilon_i$ is produced by prompting the model to articulate its reasoning before providing a final prediction, following the Chain-of-Thought prompting strategy \citep{wei_chain--thought_2022}. 

Our methodology is inspired by recent approaches that utilize LLMs as evaluators of other models' outputs \citep{dubois_alpacafarm_2023, li_self-alignment_2024, fernandes_devil_2023, bai_benchmarking_2023, saha_branch-solve-merge_2024}. This approach has shown versatility, extending beyond simple evaluation to various applications in model improvement and self-alignment strategies. For instance, researchers have employed this framework to generate self-instruct data for fine-tuning models \citep{wang_self-instruct_2023, chen_alpagasus_2023, gulcehre_reinforced_2023} and to create feedback for training reward models \citep{bai_constitutional_2022, yuan_self-rewarding_2024, wu_meta-rewarding_2024}.

For our evaluation, we employ a more capable model, \( \mathcal{M}_{\text{Judge}} \), to assess the quality of self-explanations $\varepsilon_i$ based on predefined criteria (detailed in Section \ref{sec: Quality Criteria for Effective Self-Explanations}). The evaluation process proceeds as follows:

\begin{enumerate}
    \item For each criterion \( \kappa \), \( \mathcal{M}_{\text{Judge}} \) assigns a qualitative verdict \( v_{i,\kappa} \) from the set \{excellent, satisfactory, needs improvement, unsatisfactory\}. The prompt used by \( \mathcal{M}_{\text{Judge}} \) is provided in Appendix \ref{appendix: Judge prompt}.
    \item Each verdict \( v_{i,\kappa} \) is mapped to a numerical score \( s_{i,\kappa} \) (see Appendix \ref{appendix: Judge Score Mapping}).
    \item The overall score for an explanation, \( s_i \), is computed as the sum of scores across all criteria: 
    $s_i = \sum_{k=1}^{K} s_{i,k}$
\end{enumerate}

\subsection{Pairwise Model Evaluation}
\label{sec:pairwise-model-evaluation}

To assess the quality of self-explanations generated by different models, we adopt a pairwise evaluation strategy consistent with previous work \citep{chen_alpagasus_2023, yuan_self-rewarding_2024, wu_meta-rewarding_2024}. For each input prompt \( x_i \), we generate \( N \) sample self-explanations, capturing the inherent variability in model outputs. The score for the \( n \)-th response is denoted as \( s_i^n \), with the corresponding explanation and prediction represented by the pair \( (\varepsilon_i^n, \hat{y}_i^n) \).

For a given prompt \( x_i \), we conduct \( N^2 \) pairwise comparisons between the explanations generated by two models, \( \mathcal{M}_1 \) and \( \mathcal{M}_2 \). A \textit{win} for model \( \mathcal{M}_1 \) is defined when:

\[
s_i^n(\mathcal{M}_1) > s_i^m(\mathcal{M}_2)
\]

where \( n, m \in \{1, \ldots, N\} \). The overall win rate \( W(\mathcal{M}_1, \mathcal{M}_2) \) is then calculated as follows:

\[
W(\mathcal{M}_1, \mathcal{M}_2) = \frac{1}{|\mathcal{X}|} \sum_{x_i \in \mathcal{X}} \left(\frac{1}{N^2} \sum_{n=1}^N \sum_{m=1}^N \mathbb{1}[s_i^n(\mathcal{M}_1) > s_i^m(\mathcal{M}_2)] \right)
\]

Here, \( \mathcal{X} \) denotes the set of all prompts, while \( \mathbb{1}[\cdot] \) represents the indicator function that returns 1 if the condition is true and 0 otherwise. This approach facilitates a nuanced comparison of model performance by taking into account the distribution of explanation qualities, rather than relying solely on single-point estimates. The rates for ties, defined as \( s_i^n(\mathcal{M}_1) = s_i^m(\mathcal{M}_2) \), and losses, defined as \( s_i^n(\mathcal{M}_1) < s_i^m(\mathcal{M}_2) \), are computed in a similar manner (see Appendix \ref{sec:pairwise-model-evaluation}). Throughout the evaluations presented in this work, \( \mathcal{M}_2 \) refers to the baseline model \( \mathcal{M}_{\text{base}} \).

\section{Self-Explanation Alignment with Anchor Preference Pairs}

%Building on prior work \citep{bai_constitutional_2022, wang_self-instruct_2023, yuan_self-rewarding_2024, wu_meta-rewarding_2024}, we propose a tailored methodology to fine-tune large language models (LLMs) for downstream classification tasks while preserving the quality of their self-explanations, even without annotated explanation datasets. Our approach differs from previous methods in two key ways: first, we introduce a comprehensive framework for evaluating explanation quality, focusing on the model's ability to communicate its reasoning clearly. Second, we propose a novel method, \textit{Alignment with Anchor Preference Pairs}, which constructs high-quality preference pairs in self-instruct datasets by leveraging the model’s behavior on each prompt. The methodology consists of the following steps:

In this section, we introduce a methodology for alignment designed to enhance the ability of large language models (LLMs) to articulate their reasoning—\textit{self-explanation}—even in the absence of annotated rationale explanations. However, we assume access to human-annotated data in the form of classification datasets for domain-specific adaptation, reflecting a common constraint in real-world applications, where comprehensive explanation data is often scarce or prohibitively expensive compared to classification datasets.

Building on prior work \citep{bai_constitutional_2022, wang_self-instruct_2023, yuan_self-rewarding_2024, wu_meta-rewarding_2024}, our alignment methodology incorporates familiar components such as self-instruction dataset generation, human-free evaluation of candidate responses using \textit{LLM-as-Judge}, preference pair selection, and model alignment.

However, our approach differs from previous methods in two key ways: First, for the assessment of candidate responses, we use the evaluation explanation quality framework introduced in Sections \ref{sec: Quality Criteria for Effective Self-Explanations} and \ref{sec:Self-Explanations Evaluation Methodology}. Second, we propose a novel technique, \textit{Alignment with Anchor Preference Pairs}, which improves preference pair selection by categorizing model outputs into three groups: consistently correct, consistently incorrect, and variable. By applying tailored strategies to each category, we enhance the effectiveness of DPO.

The steps of the methodology are as follows:

\begin{enumerate}
    \item Supervised fine-tuning of the base model $\mathcal{M}_{\text{Base}}$  specifically on a target classification task, resulting in $\mathcal{M}_{\text{SFT}}$.
    \item Instruct $\mathcal{M}_{\text{SFT}}$ to generate multiple explanation-prediction pairs for each prompt, and evaluate the quality of these self-explanations using the methodology outlined in Sections \ref{sec: Quality Criteria for Effective Self-Explanations} and \ref{sec:Self-Explanations Evaluation Methodology}. During alignment, the base model $\mathcal{M}_{\text{Base}}$ acts as the judge $\mathcal{M}_{\text{Judge}}$, ensuring the process remains self-contained.
    \item Construct an alignment dataset by selecting preference pairs using an anchor-based strategy (see Section \ref{sec:Preference Pairs via Anchor Selection}).
    \item Align $\mathcal{M}_{\text{SFT}}$ via DPO with the dataset created in the third step, producing the aligned model $\mathcal{M}_{\text{Anchor}}$.
\end{enumerate}

\subsection{Supervised Fine-Tuning without Annotated Explanations}
\label{sec: Supervised Fine-Tuning without Annotated Explanations}

We fine-tuned the base model, $\mathcal{M}_{\text{Base}}$, on classification datasets (the primary task) to obtain $\mathcal{M}_{\text{SFT}}$, simulating scenarios where explanation annotations are unavailable. To replicate typical domain-specific adaptations and avoid potential gains from multi-task learning, we fine-tuned a separate model for each task. During fine-tuning, loss was calculated only on the target tokens corresponding to the correct choice sentence, excluding the system instruction and question. We generated the full text of the selected option to provide richer context and preserve the model's text generation capabilities. Details on datasets and training setups are provided in Section \ref{sec: Experimental Setup}.

\subsection{Self-Instruction Creation}

We generate self-instruct data for alignment as follows:

\begin{enumerate}

    \item \textbf{Generate candidate responses:} We sample $N$ diverse pairs of explanations and predictions from $\mathcal{M}_{\text{SFT}}$, denoted as $\{\varepsilon_i^n, \hat{y}_i^n\}_{n=1}^{N}$, where $\varepsilon_i^n$ represents the explanation for the $n$-th prediction $\hat{y}_i^n$ corresponding to the prompt $x_i$.

    \item \textbf{Evaluate candidate responses:} We use the methodology described in Section \ref{sec:Self-Explanations Evaluation Methodology} to evaluate the self-explanations generated from the candidate responses, assigning a score $s_{i}^{n}$ to each explanation $\varepsilon_i^n$. During the creation of the self-instruct dataset, we employ $\mathcal{M}_{\text{base}}$ as the judge ($\mathcal{M}_{\text{Judge}}$). This ensures that the model alignment process remains self-contained, without the need for external models, except for evaluation purposes.

\end{enumerate}

\subsection{Preference Pairs via Anchor Selection}
\label{sec:Preference Pairs via Anchor Selection}

We introduce a method to enhance the selection of preference pairs by categorizing model responses into three groups: consistently correct, consistently incorrect, and variable. For each category, we apply specific strategies to construct preference pairs, which are then used in during the DPO phase. To evaluate the model's consistency on a given input prompt, a ground truth reference, or \textit{anchor}, is required. We use a classification task as the probing mechanism.

\textbf{Preference Pairs for Consistently Correct Prompts}: For input prompts $x_i$ where $\mathcal{M}_{\text{SFT}}$ consistently produces correct answers (i.e., $\hat{y}_{i}^{n} = y_{i}$ for all $n \in \{1, \ldots, N\}$), preference pairs are constructed based on the quality of the explanations. Let $s_{i}^{n}$ denote the score assigned by the judge $\mathcal{M}_{\text{Judge}}$ to the $n$-th explanation $\varepsilon_{i}^{n}$ for prompt $x_i$. We define two sets: $\mathbb{A}_{i}^{w} = \{\varepsilon_{i}^{n} : s_{i}^{n} = \max_{j \in \{1, \ldots, N\}} s_{i}^{j}\}$, which contains all explanations that achieve the highest score for prompt $x_i$, and $\mathbb{A}_{i}^{l} = \{\varepsilon_{i}^{n} : s_{i}^{n} < \max_{j \in \{1, \ldots, N\}} s_{i}^{j}\}$, which includes all explanations with scores lower than the maximum for prompt $x_i$. 

\textbf{Preference Pairs for Variable Performance}: For input prompts $x_i$ where $\mathcal{M}_{\text{SFT}}$ produces a mix of correct and incorrect predictions (i.e., $\hat{y}_{i}^{n} \neq y_{i}$ for some $n \in \{1, \ldots, N\}$), preference pairs are constructed contrastively. We define the set $\mathbb{B}_{i}^{w} = \{\varepsilon_{i}^{n} : \hat{y}_{i}^{n} = y_{i}\}$, which contains explanations associated with correct predictions. From this set, we extract $\mathbb{A}_{i}^{w} \subseteq \mathbb{B}_{i}^{w}$, the subset of explanations with the highest scores assigned by $\mathcal{M}_{\text{Judge}}$, i.e., $\mathbb{A}_{i}^{w} = \{\varepsilon_{i}^{n} \in \mathbb{B}_{i}^{w} : s_{i}^{n} = \max_{j \in \mathbb{B}_{i}^{w}} s_{i}^{j}\}$. The set $\mathbb{A}_{i}^{l} = \{\varepsilon_{i}^{n} : \hat{y}_{i}^{n} \neq y_{i} \text{ and } s_{i}^{n} < \max_{j \in \mathbb{A}_{i}^{w}} s_{i}^{j}\}$ contains explanations corresponding to incorrect predictions, with scores lower than the maximum score in $\mathbb{A}_{i}^{w}$. 

\textbf{Preference Pairs for Consistently Incorrect Prompts}: For prompts where all predictions from $\mathcal{M}_{\text{SFT}}$ are incorrect (i.e., $\hat{y}_{i}^{n} \neq y_{i}$ for all $n \in \{1, \ldots, N\}$), all corresponding explanations are placed in the set $\mathbb{A}_{i}^{l}$. To generate a winning explanation, we employ the $\mathcal{M}_{\text{Base}}$ model in a consultant role, similar to the LLM-as-a-Debater approach proposed by \citet{khan_debating_2024}. Since the inference hyperparameters for the LLM in this consulting role might differ from those used during the generation of preference pairs, we refer to this model as $\mathcal{M}_{\text{Debater}}$ to avoid confusion. Specifically, we provide the correct answer $y_{i}$ to the LLM and request an argument supporting this answer, which is then assigned to the set $\mathbb{A}_{i}^{w}$ as the winning explanation.

Finally, preference pairs are constructed for each instruction prompt $x_i$ by randomly sampling $\varepsilon_{i}^{w}$ from $\mathbb{A}_{i}^{w}$ as the winning explanation and $\varepsilon_{i}^{l}$ from $\mathbb{A}_{i}^{l}$ as the losing explanation. The resulting preference pair is denoted as $(x_i, \varepsilon_{i}^{w}, \varepsilon_{i}^{l})$. The detailed algorithm is presented in Algorithm \ref{alg:preference_pairs}.

\begin{algorithm}
    % I have added the math notation alternatives for the case statements  chose as you please :)
    \caption{Generating Preference Pairs Via Anchor Selection}
    \label{alg:preference_pairs}
    \begin{algorithmic}[1]
        \State \textbf{Input:} Instruction prompt $x_i$, model predictions $\{\hat{y}_{i}^{n}\}_{n=1}^{N}$, true label $y_i$, judge model $\mathcal{M}_{\text{Judge}}$, debater model $\mathcal{M}_{\text{Debater}}$
        \State \textbf{Output:} Preference pairs $(x_i, \varepsilon_{i}^{w}, \varepsilon_{i}^{l})$
        
        \State \textbf{Initialize:} $\mathbb{A}_{i}^{w} \gets \emptyset$, $\mathbb{A}_{i}^{l} \gets \emptyset$
        
        \If{$\hat{y}_{i}^{n} = y_i$ for all $n \in \{1, \ldots, N\}$} 
        \Comment{Consistently Correct Prompts}
        % \If{$\hat{y}_{i}^{n} = y_i, \forall n \in \{1, \ldots, N\}$} 
            \For{each explanation $\varepsilon_{i}^{n}$}
                \State Compute score $s_{i}^{n}$ from $\mathcal{M}_{\text{Judge}}$
            \EndFor
            \State $\mathbb{A}_{i}^{w} \gets \{\varepsilon_{i}^{n} : s_{i}^{n} = \max_{j \in \{1, \ldots, N\}} s_{i}^{j}\}$
            
            \State $\mathbb{A}_{i}^{l} \gets \{\varepsilon_{i}^{n} : s_{i}^{n} = \min_{j \in \{1, \ldots, N\}} s_{i}^{j}\}$

        \ElsIf{$\hat{y}_{i}^{n} \neq y_i$ for some $n \in \{1, \ldots, N\}$} \Comment{Variable Performance Prompts}
            \State $\mathbb{B}_{i}^{w} \gets \{\varepsilon_{i}^{n} : \hat{y}_{i}^{n} = y_i\}$
            \State $\mathbb{A}_{i}^{w} \gets \{\varepsilon_{i}^{n} \in \mathbb{B}_{i}^{w} : s_{i}^{n} = \max_{j \in \mathbb{B}_{i}^{w}} s_{i}^{j}\}$
            \State $\mathbb{A}_{i}^{l} \gets \{\varepsilon_{i}^{n} : \hat{y}_{i}^{n} \neq y_i \land s_{i}^{n} < \max_{j \in \mathbb{A}_{i}^{w}} s_{i}^{j}\}$
        
        \Else \Comment{Consistently Incorrect Prompts}
            % \State $\mathbb{A}_{i}^{l} \gets \{\varepsilon_{i}^{n} : \hat{y}_{i}^{n} \neq y_i, \forall n \in \{1, \ldots, N\}\}$
            \State $\mathbb{A}_{i}^{l} \gets \{\varepsilon_{i}^{n} : \hat{y}_{i}^{n} \neq y_i \text{ for all } n \in \{1, \ldots, N\}\}$
            \State Generate argument $\varepsilon_i^{\text{\text{debater}}}$ using $\mathcal{M}_{\text{Debater}}$ given $y_i$
            \State $\mathbb{A}_{i}^{w} \gets \{\varepsilon_i^{\text{debater}}\}$
        \EndIf
        
        \State \textbf{Sample} $\varepsilon_{i}^{w}$ from $\mathbb{A}_{i}^{w}$
        \State \textbf{Sample} $\varepsilon_{i}^{l}$ from $\mathbb{A}_{i}^{l}$
        % drop if $s_{i}^{w}$  == $\s_{i}^{l}$
        \State \textbf{Return} $(x_i, \varepsilon_{i}^{w}, \varepsilon_{i}^{l})$
    \end{algorithmic}
\end{algorithm}

\section{Experiments}

In all experiments, we utilized \texttt{Llama-3-8B-Instruct} as our base model. Our study involved comparing four distinct model configurations:

\begin{enumerate}
    \item $\mathcal{M}_{\text{Base}}$: The base model, which remains unmodified.
    \item $\mathcal{M}_{\text{SFT}}$: This model was obtained by performing supervised fine-tuning on the $\mathcal{M}_{\text{Base}}$ model  using only classification tasks, simulating scenarios where explanation annotations are not available. 
    \item $\mathcal{M}_{\text{Rank}}$: The $\mathcal{M}_{\text{SFT}}$ model was further refined using DPO, employing a self-instruct dataset constructed from rank-ordered preference pairs derived solely from judge-based evaluations of the explanations. This approach aligns with methodologies described in \cite{bai_constitutional_2022, wang_self-instruct_2023, yuan_self-rewarding_2024, wu_meta-rewarding_2024}.
    \item $\mathcal{M}_{\text{Anchor}}$ (ours): Similar to $\mathcal{M}_{\text{Rank}}$, this model was refined using DPO but utilized a self-instruct dataset created with our proposed anchored method for selecting preference pairs.
\end{enumerate}

Since both $\mathcal{M}_{\text{Rank}}$ and $\mathcal{M}_{\text{Anchor}}$ undergo an additional stage of DPO alignment with the self-instruct dataset, we will collectively refer to them as self-aligned models in comparison to $\mathcal{M}_{\text{SFT}}$.

\subsection{Experimental Setup}
\label{sec: Experimental Setup}

\textbf{Datasets}: We selected four datasets for our experiments:  \texttt{AQuA-Rat} \citep{ling_program_2017}, \texttt{ARC-Challenge} \citep{clark_think_2018},  \texttt{LogiQA} \citep{liu_logiqa_2020}, and \texttt{OpenbookQA} \citep{mihaylov_can_2018}. These datasets are established benchmarks for reasoning tasks, requiring a challenging reasoning process, which makes them an ideal fit for evaluating the quality of self-explanations. A key factor in their selection was the size of their training sets, which provided a sufficient number of input prompts to support the creation of the self-instruction dataset. In the case of \texttt{AQuA-Rat}, we sampled 5,000 examples due to computational constraints. For evaluation, we used the test split of each dataset.

\textbf{SFT Training Details}: 
For \( \mathcal{M}_{\text{SFT}} \), we used the AdamW optimizer with a learning rate of \( 5 \times 10^{-5} \) for one epoch, following a cosine schedule with 10\% warmup steps. Gradient clipping was set to 0.3, and we used an effective batch size of 12. Loss was computed only on the assistant's completions. Instead of fine-tuning the entire model, we applied a LoRA adapter (\( \alpha = 128 \), dropout = 0.05, rank \( r = 256 \)) to all linear layers. LoRA adapters were used to accelerate training and to act as a regularization method \citep{biderman_lora_2024}, addressing the overfitting tendencies of DPO \citep{thakkar_deep_2024}, which is applied during the later alignment phase.

\textbf{Self-Instruct Dataset}: To ensure the integrity of our evaluation process, we created separate self-instruct datasets for each benchmark. These datasets were built using input prompts specific to each task, ensuring that the DPO alignment data remained uncontaminated by exposure to multiple tasks. This approach prevents the artificial inflation of results that could occur if models were aligned across diverse tasks, unlike SFT models, which are trained on one classification task at a time.
To create the self-instruct dataset for aligning $\mathcal{M}_{\text{Rank}}$ and $\mathcal{M}_{\text{Anchor}}$, we sampled $N=4$ responses from $\mathcal{M}_{\text{SFT}}$ for each input prompt (settings: temperature $T=0.6$ and top-k value of $0.9$). This sample size provided a reasonable assessment of the model's consistency and variability. The prompt used is presented in Appendix \ref{appendix: Prompts inference per task}. In cases of consistently incorrect responses, $\mathcal{M}_{\text{Base}}$ was employed as the debater model (\texttt{Llama-3-8B-Instruct}) with adjusted parameters ($T=0.5$, top-k $0.9$). The responses were scored by $\mathcal{M}_{\text{Judge}}$, which was the same base model, ensuring a self-contained alignment process. This setup differs from the evaluation phase, where a more capable model serves as the judge. The scoring of explanations followed the methodology outlined in Section \ref{sec:Self-Explanations Evaluation Methodology}, with $\mathcal{M}_{\text{Judge}}$ using fixed inference parameters ($T=0$). For $\mathcal{M}_{\text{Rank}}$, preference pairs were chosen based on the assigned scores, with the highest-scoring explanation designated as the winner, and the losing explanation randomly selected from the remaining candidates. The preference pairs for $\mathcal{M}_{\text{Anchor}}$ were selected using the methodology outlined in Section \ref{sec:Preference Pairs via Anchor Selection}, and these pairs were then used to align the models through DPO.

\textbf{DPO Training Details}: 
For DPO-aligned models ($\mathcal{M}_{\text{Rank}}$, $\mathcal{M}_{\text{Anchor}}$), we used similar hyperparameters as in the SFT phase but reduced the learning rate to \( 5 \times 10^{-7} \) and trained for 2.6k steps with an effective batch size of 6. The DPO process used a $\beta$ value of 0.1 and updated the LoRA weights obtained during SFT.

\textbf{Evaluation}: We evaluated our models along two main dimensions: prediction accuracy and self-explanation quality. To account for variability in model outputs, we generated $N=16$ explanation-prediction pairs per input prompt. The inference settings mirrored those used to create the self-instruction dataset, with a temperature of $T=0.6$ and top-k set to $0.9$ and the same prompt used during the creation of the self-instruct dataset (see Appendix \ref{appendix: CoT prompt for LogiQA task}). Average prediction accuracy was used to measure performance on downstream tasks. To assess self-explanation quality, we performed head-to-head comparisons between the aligned models ($\mathcal{M}_{\text{SFT}}$, $\mathcal{M}_{\text{Rank}}$, $\mathcal{M}_{\text{Anchor}}$) and the base model ($\mathcal{M}_{\text{Base}}$). These comparisons followed the methodology outlined in Section \ref{sec:Self-Explanations Evaluation Methodology}, employing \texttt{Llama-3-70B-Instruct} as $\mathcal{M}_{\text{Judge}}$.

\subsection{Results}
Table \ref{results} reports the average classification accuracy for each model, along with pairwise comparisons of self-explanation quality across multiple benchmark datasets. The win, tie, and loss rates are calculated by comparing the aligned models against \( \mathcal{M}_{\text{Base}} \).

\begin{table}[h]
\caption{\textbf{Comparison of Aligned Models}. Average accuracy  is presented along with head-to-head comparisons for self-explanation quality. The results show that $\mathcal{M}_{\text{Anchor}}$ achieves comparable or superior accuracy and attains the highest win rate when compared to $\mathcal{M}_{\text{SFT}}$ and $\mathcal{M}_{\text{Rank}}$.}

\label{results}
\begin{center}
\begin{tabular}{lccccccc}
\toprule 
\multirow{2}{*}{\textbf{Dataset}} & \multirow{2}{*}{\textbf{\begin{tabular}{c}$\mathcal{M}_{\text{Base}}$ \\ Acc. (\%)\end{tabular}}} & \multirow{2}{*}{\textbf{\begin{tabular}{c}$\mathcal{M}_{\text{Align}}$ \\ Type\end{tabular}}} & \multirow{2}{*}{\textbf{\begin{tabular}{c}$\mathcal{M}_{\text{Align}}$\\ Acc. (\%)\end{tabular}}} & \multicolumn{3}{c}{\textbf{$\varepsilon$ Win Rate Eval. (\%)}} & \multirow{2}{*}{$\Delta_{W-L}$ } \\
\cmidrule(lr){5-7}
& & & &\textbf{Win $\uparrow$} & \textbf{Tie} & \textbf{Loss $\downarrow$} & \\
\midrule \midrule

& & $\mathcal{M}_{\text{SFT}}$ & $47.7_{\pm 2.7}$ & $11.3$ & $61.9$ & $26.9$ & -15.6 \\
AQuA Rat & $47.1_{\pm 2.9}$ & $\mathcal{M}_{\text{Rank}}$ & $48.3_{\pm 2.1}$ & $11.5$ & $63.1$ & $25.4$ & -13.9 \\
& &  $\mathcal{M}_{\text{Anchor}}$ & \cellcolor{lightgray}$51.1_{\pm 3.0}$ & \cellcolor{lightgray}$12.6$ & \cellcolor{lightgray}$70.8$ & \cellcolor{lightgray}$16.7$ & \cellcolor{lightgray} -4.1 \\
\cmidrule(lr){3-8}

& & $\mathcal{M}_{\text{SFT}}$ & $81.0_{\pm 0.7}$ & $9.6$ & $54.4$ & $36.0$ & -26.4 \\
ARC-Challenge & $76.4_{\pm 0.7}$  & $\mathcal{M}_{\text{Rank}}$ & $81.9_{\pm 1.1}$ & $17.72$ & $61.4$ & $20.9$ & -3.18 \\
& & $\mathcal{M}_{\text{Anchor}}$  &\cellcolor{lightgray}$82.0_{\pm 0.9}$ & \cellcolor{lightgray}$21.5$ & \cellcolor{lightgray}$60.3$ & \cellcolor{lightgray}$18.2$ & \cellcolor{lightgray} 3.3 \\
\cmidrule(lr){3-8}

& & $\mathcal{M}_{\text{SFT}}$ & $45.2_{\pm 0.7}$ & $14.7$ & $39.8$ & $45.6$ & -30.9 \\
LogiQA & $41.4_{\pm 1.1}$ & $\mathcal{M}_{\text{Rank}}$ & $46.0_{\pm 1.5}$ & $22.0$ & $50.1$ & $27.9$ & -5.9 \\
& &  $\mathcal{M}_{\text{Anchor}}$ & \cellcolor{lightgray}$46.6_{\pm 2.2}$ & \cellcolor{lightgray}$26.6$ & \cellcolor{lightgray}$53.8$ & \cellcolor{lightgray}$19.7$ & \cellcolor{lightgray} 6.9 \\
\cmidrule(lr){3-8}

& & $\mathcal{M}_{\text{SFT}}$ & \cellcolor{lightgray}$87.4_{\pm 1.1}$ & $11.3$ & $54.0$ & $34.6$ & -23.3 \\
OpenbookQA & $71.7_{\pm 1.3}$ & $\mathcal{M}_{\text{Rank}}$ & $87.0_{\pm 1.1}$ & $15.4$ & $60.1$ & $24.5$ & -9.1 \\
& & $\mathcal{M}_{\text{Anchor}}$  & $87.0_{\pm 0.9}$ & \cellcolor{lightgray}$16.7$ & \cellcolor{lightgray}$59.6$ & \cellcolor{lightgray}$23.7$ & \cellcolor{lightgray}  -7 \\
\bottomrule
\end{tabular}
\end{center}
\end{table}

\subsubsection{Impact of Supervised Fine-Tuning on Self-Explanations} \label{sec:Supervised_Fine_Tuning_on_Self-Explanations}

We observed a significant trade-off in evaluation results before and after applying supervised fine-tuning on a classification task (see Table \ref{results}). While SFT notably improved classification accuracy, it resulted in a substantial decline in the quality of self-explanations compared to the base model. The decline in explanation quality, as measured by the win-loss rate difference (\(\Delta_{W-L}\)), ranged from 15.6\% to 30.9\% across benchmarks.

Building on evidence that supervised fine-tuning can improve performance on specific tasks at the expense of a model's generalization abilities \citep{yang_unveiling_2024, kirk_understanding_2024}, we hypothesize that this decline occurs because the task of selecting predefined answers does not inherently encourage the model to articulate its reasoning, leading to a specialization that diminishes the quality of the generated explanations.

These findings highlight the necessity for alignment techniques that can preserve high-quality explanations in situations where datasets with annotated explanations are unavailable for fine-tuning.

\subsubsection{Analysis of Self-Aligned Models} \label{sec:comparison_aligned_models}

\textbf{Prediction Accuracy}: Our experiments demonstrate that the self-aligned models, $\mathcal{M}_{\text{Rank}}$ and $\mathcal{M}_{\text{Anchor}}$, maintain the classification accuracy improvements achieved by the seed model, $\mathcal{M}_{\text{SFT}}$, over the base model, $\mathcal{M}_{\text{Base}}$ (see Table \ref{results}). Notably, $\mathcal{M}_{\text{Anchor}}$ consistently achieves the highest, or at least comparable, accuracy across all datasets. We believe that this improvement can be attributed to the fact that the model’s predictions ($\hat{y}_i$) are compared to the ground truth ($y_i$) to determine how to treat the self-explanation \((e_i\)) during the selection of preference pairs while employing the anchor strategy, thereby providing a more informative learning signal.

\textbf{Self-Explanation Quality}: Pairwise evaluations of self-explanation quality (see Table \ref{results}) indicate that the initial decline in explanation performance observed in $\mathcal{M}_{\text{SFT}}$ is partially inherited by both $\mathcal{M}_{\text{Rank}}$ and $\mathcal{M}_{\text{Anchor}}$, as they use $\mathcal{M}_{\text{SFT}}$ as the seed model during the DPO alignment phase. Nevertheless, both $\mathcal{M}_{\text{Rank}}$ and $\mathcal{M}_{\text{Anchor}}$ demonstrate significant improvements in explanation quality compared to $\mathcal{M}_{\text{SFT}}$, with $\mathcal{M}_{\text{Anchor}}$ exhibiting the strongest performance. Compared to the base model, $\mathcal{M}_{\text{Anchor}}$ also shows positive $\Delta_{W-L}$ margins on ARC-Challenge (3.3\%) and LogiQA (6.9\%), and it significantly narrows the gap in explanation quality introduced by SFT across the remaining benchmark datasets.

\begin{figure}[h]
\begin{center}
%\framebox[4.0in]{$\;$}
%\fbox{\rule[-.5cm]{0cm}{4cm} \rule[-.5cm]{4cm}{0cm}}
\includegraphics[width=\textwidth]{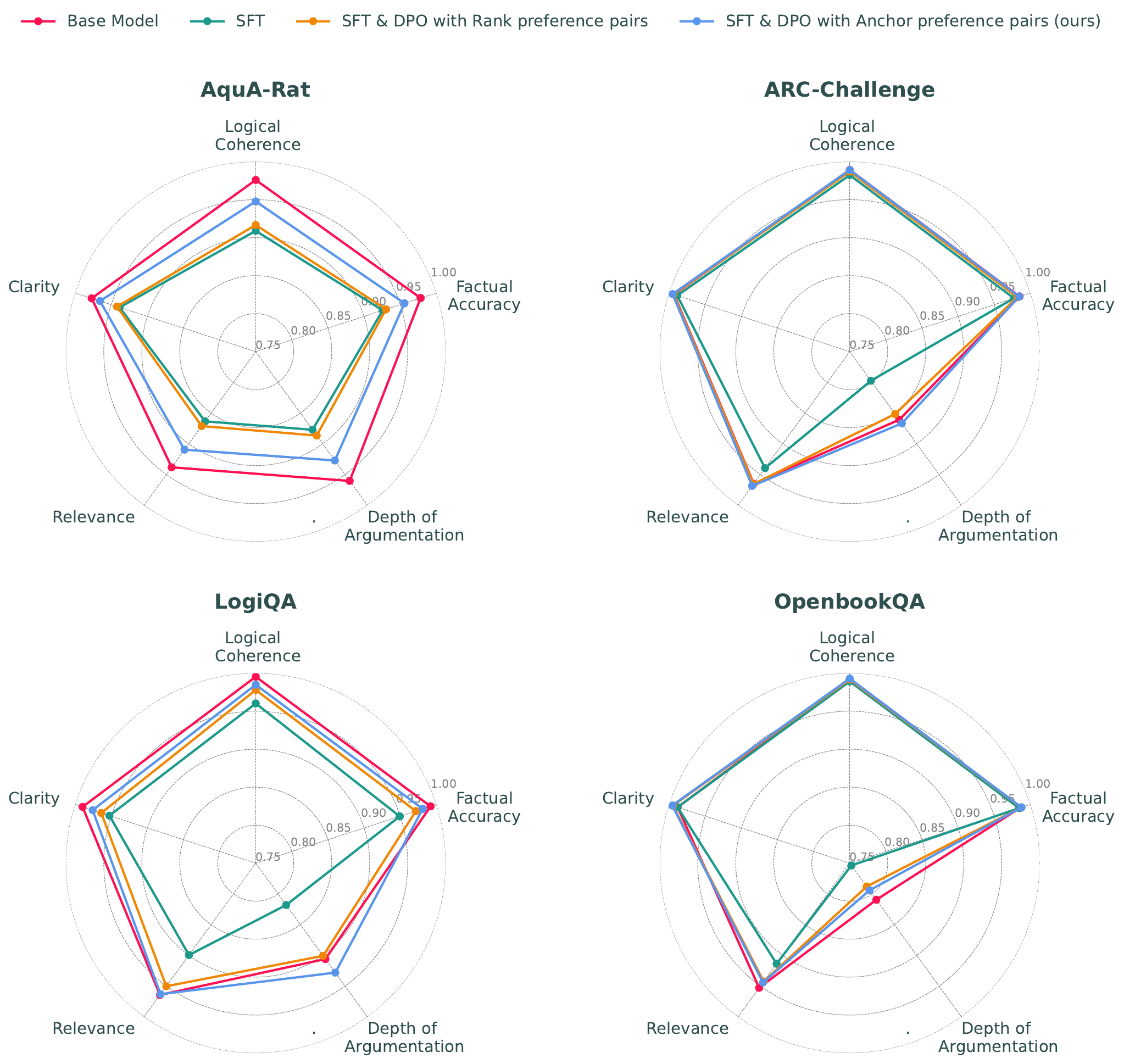} % Adjust width as needed
\end{center}

\caption{\textbf{Average Self-Explanation Scores per Evaluation Criterion}. Average scores for each evaluation criterion used to assess self-explanations, as described in Section \ref{sec: Quality Criteria for Effective Self-Explanations}. The scores are provided for all evaluated models across the benchmark datasets.}

\label{fig: Results self-explanation per criteria}

\end{figure}

\subsubsection{Analysis of Individual Evaluation Dimensions}

Figure \ref{fig: Results self-explanation per criteria} presents the average scores for each evaluation criterion used to assess self-explanations, as described in Section \ref{sec: Quality Criteria for Effective Self-Explanations}, for all evaluated models across the benchmark datasets.

Overall, the self-aligned models outperform $\mathcal{M}_{\text{SFT}}$ across all evaluation criteria, with $\mathcal{M}_{\text{Anchor}}$ consistently achieving better results than $\mathcal{M}_{\text{Rank}}$.

Additionally, we observe that the degradation in self-explanation quality due to SFT varies significantly depending on the dataset used for fine-tuning. Two notable trends emerge from the analysis. First, for more complex tasks—where complexity is measured by lower test accuracy—such as \texttt{AQuA-Rat} and \texttt{LogiQA}, the decline in explanation quality is more pronounced across all criteria. Second, evaluation dimensions for which the base model originally received lower scores tend to experience a more significant drop in performance after SFT.

\subsubsection{Impact of Preference Pairs Category Distribution}

\begin{figure}[h]
\begin{center}
%\framebox[4.0in]{$\;$}
%\fbox{\rule[-.5cm]{0cm}{4cm} \rule[-.5cm]{4cm}{0cm}}
\includegraphics[width= 
\textwidth]{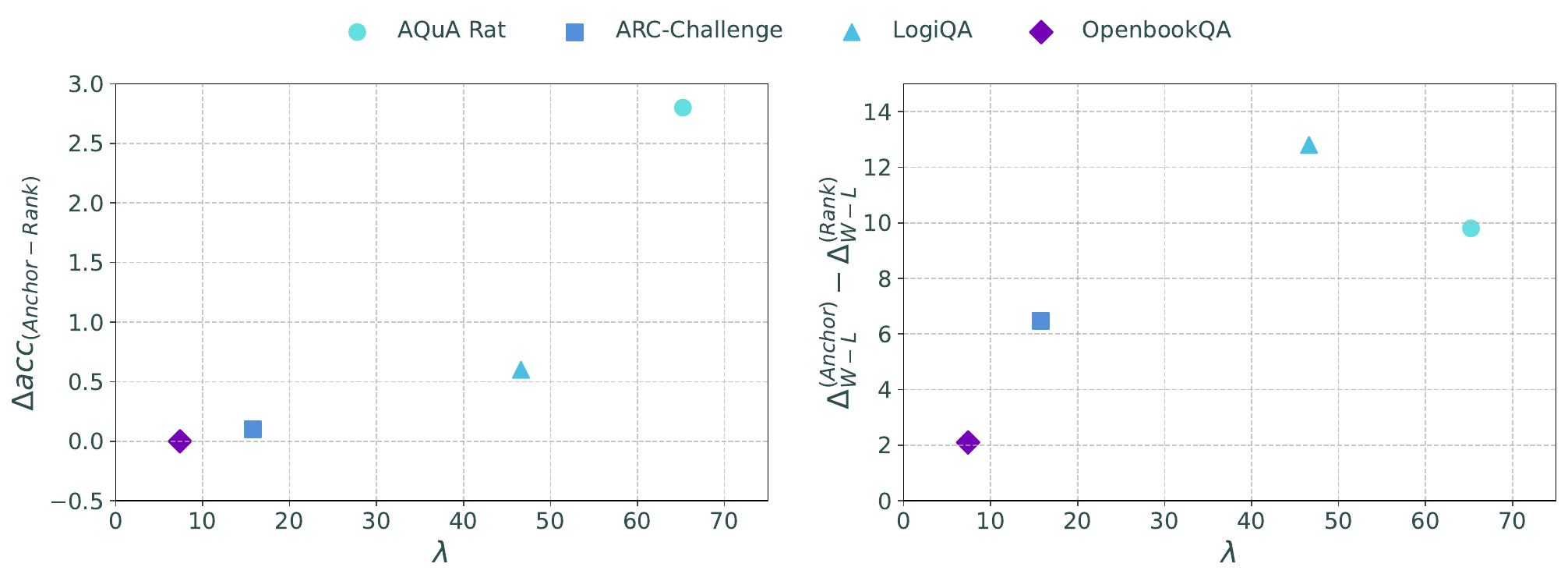} % Adjust width as needed
\end{center}

\caption{\textbf{Impact of Preference Pairs Category Distribution:} Presents the relative improvements in accuracy (\textit{left}) and \(\Delta_{W-L}\) (\textit{right}) between \(\mathcal{M}_{\text{Anchor}}\) and \(\mathcal{M}_{\text{Rank}}\) with respect to \(\lambda\).}

\label{fig: lambda}

\end{figure}

We define \(\lambda\) as the proportion of the self-instruct dataset used to align \(\mathcal{M}_{\text{Anchor}}\), which corresponds to preference pairs selected under the \textit{consistently-incorrect} or \textit{variable} strategies (see Section \ref{sec:Preference Pairs via Anchor Selection}). This metric, \(\lambda\), provides insight into how the composition of the self-instruct dataset for \(\mathcal{M}_{\text{Anchor}}\) differs from that of \(\mathcal{M}_{\text{Rank}}\), which selects pairs based solely on scores assigned by judges, following the \textit{consistently-correct} strategy. We evaluated improvements by analyzing the differences in accuracy and \(\Delta_{W-L}\) between \(\mathcal{M}_{\text{Anchor}}\) and \(\mathcal{M}_{\text{Rank}}\) in relation to \(\lambda\) (see Figure \ref{fig: lambda}). In both cases, we observed a trend showing that \(\mathcal{M}_{\text{Anchor}}\) demonstrates a greater relative improvement compared to \(\mathcal{M}_{\text{Rank}}\) as \(\lambda\) increases. This supports our design principle that tailoring strategies based on model behavior is crucial for improving the quality of self-instruct datasets and avoiding the reinforcement of problematic behavior.

\section{Related Work}

\textbf{LLM-as-Evaluator}: This concept refers to the ability of large language models (LLMs) to evaluate the outputs of other LLMs, a technique commonly referred to as LLM-as-a-Judge. This approach has gained considerable traction in recent years \citep{dubois_alpacafarm_2023, li_self-alignment_2024, fernandes_devil_2023, bai_benchmarking_2023} and is frequently used to assess LLM performance across various downstream tasks \citep{zheng_judging_2023}. It has proven particularly effective in automating evaluations, as demonstrated on platforms like LMSys Chatbot Arena. Key implementations include direct scoring based on specific criteria, pairwise comparisons \citep{liu_aligning_2024}, reference-based evaluations, and ensemble methods \citep{verga_replacing_2024}. While LLM-as-a-Judge offers scalability and consistency, it can also inherit biases from the evaluation model, potentially amplifying problematic outputs \citep{huang_limitations_2024}. Despite these challenges, it remains a valuable tool due to its efficiency and cost-effectiveness in evaluating LLM systems. In our work, we introduce a framework for the qualitative assessment of \textit{self-explanations} using the LLM-as-a-Judge technique, designed to evaluate how effectively a model conveys its reasoning.

\textbf{Self-Alignment}: Several approaches have been developed to improve LLMs without requiring human-annotated feedback. One method involves fine-tuning models using high-quality, self-generated input-output pairs \citep{wang_self-instruct_2023, chen_alpagasus_2023, gulcehre_reinforced_2023}, though this can perpetuate biases in example selection without a clear mechanism for improving selection quality. Another influential approach is Constitutional AI \citep{bai_constitutional_2022}, where an LLM provides feedback and refines responses, which are then used to train a separate, static reward model. Building on this concept, \cite{yuan_self-rewarding_2024} and \citet{wu_meta-rewarding_2024} proposed using the LLM itself as a dynamic reward model, eliminating the need for a static one. This allows for continuous improvement in both generation and evaluation capabilities through iterative training processes. In our work, we introduce a novel method for creating a self-instruct dataset. Our approach, called \textit{Alignment with Anchor Preference Pairs}, enhances preference pair selection by categorizing model behavior in response to each input prompt and applying tailored strategies for each category. To evaluate a model’s consistency for a given input prompt, an anchor—i.e., a ground-truth reference—is required, which we derive from a classification task used as a probing mechanism.

\textbf{LLM-as-a-Debater}: This adversarial approach aims to improve model performance through argumentation. In \citet{perez_finding_2019}, debaters are limited to extracting relevant statements from a source text, rather than generating original arguments. \citet{du_improving_2023} extended this concept by involving multiple LLM instances to debate their individual responses over several rounds, eventually converging on a shared final answer. \citet{khan_debating_2024} further developed this approach by using debate-like scenarios to challenge and refine model outputs through simulated arguments. In our work, we adopt the LLM-as-a-Debater approach in the role of a consultant, specifically following \citet{khan_debating_2024}, for cases where the model’s response to certain input prompts is \textit{consistently incorrect}. This strategy enables the creation of self-instruct examples that  avoid reinforcing problematic behavior.

\section{Limitations}

We acknowledge some limitations in our approach. First, evaluating the model’s consistency on a given input prompt requires a  anchor—ground truth reference. Consequently, the selection of preference pairs via the anchor strategy relies on a classification task as the probing mechanism, which restricts its applicability. Second, when ranking the quality of self-explanations, we assign equal weights across all evaluation dimensions. This uniform weighting may not accurately reflect the varying significance of different aspects of explanation quality, which can differ depending on the user or specific application. Moreover, this approach may overlook instances where individual explanations degrade in separate criteria, potentially leading to preference pairs where score differences arise from unrelated factors.

Finally, using the base model as the judge during the creation of the self-instruct dataset eliminates the need for a more capable model but introduces a static evaluation process. As the model improves, the judge may fail to capture important evaluation nuances. Iteratively enhancing the judge’s capabilities, similar to the approaches in \cite{yuan_self-rewarding_2024} and \cite{wu_meta-rewarding_2024}, could help mitigate this issue.

\section{Conclusion}

In this work, we introduced a methodology for alignment that enhances LLMs' ability to generate high-quality self-explanations, even in the absence of annotated rationale explanations. Our approach provides an end-to-end solution for aligning LLMs on classification tasks, ensuring that they not only produce accurate predictions but also articulate coherent explanations for their decisions. This is achieved through three core components: evaluating the quality of generated explanations, creating self-instruct datasets, and aligning the model. Central to our approach is Alignment with Anchor Preference Pairs, a novel method that refines preference pair selection by categorizing model outputs into three groups: consistently correct, consistently incorrect, and variable. For each category, we apply tailored strategies to construct preference pairs, which are then used in DPO. Our empirical results demonstrate that this method consistently improves explanation quality, reducing the degradation caused by task specialization. Furthermore, we show that using anchor preference pairs outperforms self-alignment methods that rely solely on judge-based evaluations for preference pair selection.

\bibliography{iclr2025_conference}
\bibliographystyle{iclr2025_conference}

\appendix

\section{Pairwise Model Evaluation} \label{sec:pairwise-model-evaluation}

To compare the performance of two models, denoted as \( \mathcal{M}_1 \) and \( \mathcal{M}_2 \), we perform a pairwise evaluation of the self-explanations generated for a given prompt \( x_i \). Each model produces \( N \) explanations, and we compare each explanation from \( \mathcal{M}_1 \) with every explanation from \( \mathcal{M}_2 \), resulting in \( N^2 \) pairwise comparisons.

For a given comparison between the \( n \)-th explanation from model \( \mathcal{M}_1 \) and the \( m \)-th explanation from model \( \mathcal{M}_2 \), where \( n, m \in \{1, \ldots, N\} \), we compare the corresponding scores, \( s_i^n(\mathcal{M}_1) \) and \( s_i^m(\mathcal{M}_2) \). A \textit{win} for \( \mathcal{M}_1 \) is recorded if the score from \( \mathcal{M}_1 \) is strictly greater than that from \( \mathcal{M}_2 \):

\[
s_i^n(\mathcal{M}_1) > s_i^m(\mathcal{M}_2)
\]

Conversely, a \textit{loss} for \( \mathcal{M}_1 \) occurs if the score from \( \mathcal{M}_1 \) is strictly less than the score from \( \mathcal{M}_2 \):

\[
s_i^n(\mathcal{M}_1) < s_i^m(\mathcal{M}_2)
\]

A \textit{tie} is defined when both scores are equal:

\[
s_i^n(\mathcal{M}_1) = s_i^m(\mathcal{M}_2)
\]

For each prompt \( x_i \), we count the total number of wins, losses, and ties across all \( N^2 \) comparisons between the explanations from both models. To summarize the performance of the models across the entire dataset, we compute the win rate, tie rate, and loss rate.

The win rate \( W(\mathcal{M}_1, \mathcal{M}_2) \) is the average proportion of pairwise comparisons in which model \( \mathcal{M}_1 \) outperforms model \( \mathcal{M}_2 \) across all prompts in the set \( \mathcal{X} \). It is computed as:

\[
W(\mathcal{M}_1, \mathcal{M}_2) = \frac{1}{|\mathcal{X}|} \sum_{x_i \in \mathcal{X}} \left( \frac{1}{N^2} \sum_{n=1}^N \sum_{m=1}^N \mathbb{1}[s_i^n(\mathcal{M}_1) > s_i^m(\mathcal{M}_2)] \right)
\]

Here, \( \mathcal{X} \) is the set of all prompts, and \( \mathbb{1}[\cdot] \) is the indicator function, which returns 1 if the condition inside the brackets is true (i.e., if \( \mathcal{M}_1 \) wins) and 0 otherwise.

Similarly, we define the tie rate \( T(\mathcal{M}_1, \mathcal{M}_2) \) as the proportion of pairwise comparisons where the models perform equally:

\[
T(\mathcal{M}_1, \mathcal{M}_2) = \frac{1}{|\mathcal{X}|} \sum_{x_i \in \mathcal{X}} \left( \frac{1}{N^2} \sum_{n=1}^N \sum_{m=1}^N \mathbb{1}[s_i^n(\mathcal{M}_1) = s_i^m(\mathcal{M}_2)] \right)
\]

The loss rate \( L(\mathcal{M}_1, \mathcal{M}_2) \) captures the proportion of comparisons where \( \mathcal{M}_1 \) performs worse than \( \mathcal{M}_2 \):

\[
L(\mathcal{M}_1, \mathcal{M}_2) = \frac{1}{|\mathcal{X}|} \sum_{x_i \in \mathcal{X}} \left( \frac{1}{N^2} \sum_{n=1}^N \sum_{m=1}^N \mathbb{1}[s_i^n(\mathcal{M}_1) < s_i^m(\mathcal{M}_2)] \right)
\]

By evaluating the win, tie, and loss rates, we obtain a comprehensive picture of how the two models compare in terms of generating higher-quality self-explanations. This approach accounts for variability in the quality of explanations generated by the models, providing a more nuanced and detailed evaluation than methods relying solely on single-point estimates.

\section{Inference Parameters}
Table \ref{tab:training-params} summarizes the inference parameters, including temperature and top-k, used for each component, such as the judge, consultant, and sampler.

\begin{table}[H]
\caption{\textbf{Inference parameters per component}}
\label{tab:training-params}
\begin{center}
\begin{tabular}{lcc}
\toprule
\textbf{Component} & $\textbf{Temperature}$ & $\textbf{Top-k}$ \\
\midrule
\midrule
Judge & $0.0$ & \\
Consultant & $0.5$ & $0.9$ \\
Sampler & $0.6$ & $0.9$ \\
\bottomrule
\end{tabular}
\end{center}
\end{table}

\section{Judge Component}

The judge model \( \mathcal{M}_{\text{Judge}} \) evaluates the quality of self-explanation, denoted as \( \varepsilon_i \), associated with an input prompt \( x_i \). based on predefined criteria, which are elaborated in Section \ref{sec: Quality Criteria for Effective Self-Explanations}. The evaluation process proceed as follows:

\begin{enumerate}
    \item For each criterion \( \kappa \), \( \mathcal{M}_{\text{Judge}} \) assigns a qualitative verdict \( v_{i,\kappa} \) from the set \{excellent, satisfactory, needs improvement, unsatisfactory\}. The prompt used by \( \mathcal{M}_{\text{Judge}} \) is provided in Appendix \ref{appendix: Judge prompt}.
    \[
    \mathcal{M}_{\text{Judge}}(x_i, \varepsilon_i) \rightarrow \{ v_{i,\kappa} \} \text{ for } \kappa \in \{1, \dots, K\}
    \]
    \item Each verdict \( v_{i,\kappa} \) is mapped to a numerical score \( s_{i,\kappa} \) (see Appendix \ref{appendix: Judge Score Mapping}).
    \item The overall score for an explanation, \( s_i \), is computed as the sum of scores across all criteria: 
    \[
    s_i = \sum_{k=1}^{K} s_{i,k}
    \]
\end{enumerate}

\subsection{Judge Score Mapping}
\label{appendix: Judge Score Mapping}

Each verdict \( v_{i,\kappa} \), assigned by \( \mathcal{M}_{\text{Judge}} \) for criterion \( \kappa \) on self-explanation \( \varepsilon_i \) corresponding to prompt \( x_i \), is mapped to a numerical score \( s_{i,\kappa} \) as follows:

\[
s_{i,\kappa} =
\begin{cases}
1.0 & \text{if } v_{i,\kappa} = \textbf{\texttt{Excellent}}, \\
0.8 & \text{if } v_{i,\kappa} = \textbf{\texttt{Good}}, \\
0.6 & \text{if } v_{i,\kappa} = \textbf{\texttt{Satisfactory}}, \\
0.2 & \text{if } v_{i,\kappa} = \textbf{\texttt{Needs Improvement}}, \\
0.0 & \text{if } v_{i,\kappa} = \textbf{\texttt{Unsatisfactory}}.
\end{cases}
\]

In this mapping, scores \( s_{i,\kappa} \in \{1.0, 0.8, 0.6, 0.2, 0.0\} \) represent the numerical evaluation of the verdict, with higher values indicating better performance.

\subsection{Judge prompt}
\label{appendix: Judge prompt}
Table \ref{tab: Judge prompt} presents the complete prompt that guides the judge model in evaluating the self-explanations.

\begin{table}[H]
\caption{\textbf{Judge prompt}.}
\label{tab: Judge prompt}
\begin{center}
\begin{tabular}{p{13cm}} % Adjust the width as needed
\toprule[0.95pt]
\textbf{Prompt Judge}\\
\midrule[0.95pt]
\midrule[0.95pt]
%\rowcolor{lightgray} 
\scriptsize

\texttt{\textbf{System}: You are an expert judge tasked with evaluating the quality and correctness of a STATEMENT in response to a given QUESTION. 
\newline
\newline
Your goal is to assess whether the STATEMENT accurately and comprehensively answers the QUESTION while providing sound reasoning and clear explanations. Be vigilant for any errors, misleading information, or gaps in the reasoning.
\newline}

\texttt{Evaluate the STATEMENT based on the following criteria:}
\begin{itemize}
    \item \texttt{\textbf{Factual Accuracy:} Are all specific claims, data points, and facts in the STATEMENT verifiably correct?}
    \item \texttt{\textbf{Logical Coherence:} Does the STATEMENT follow a clear, logical sequence of reasoning without contradictions, and do all parts logically connect to form a cohesive and unified explanation?}
    \item \texttt{\textbf{Clarity and Comprehensibility:} Is the STATEMENT expressed clearly and precisely, using appropriate terminology and effectively explaining complex ideas?}
    \item \texttt{\textbf{Relevance and Completeness:} Does the STATEMENT fully address all relevant aspects of the QUESTION, ensuring that it directly responds to the specific context or requirements, leaving no key details omitted?}
    \item \texttt{\textbf{Depth of Argumentation:} Does the STATEMENT provide strong reasoning and credible evidence to support its conclusions?}
\end{itemize}

\texttt{
\newline
For each criterion, select one of the following assessments:}
\begin{itemize}
    \item \texttt{\textbf{EXCELLENT:} The STATEMENT exemplifies this criterion, setting a high standard.}
    \item \texttt{\textbf{GOOD:} The STATEMENT meets this criterion well, with only minor room for improvement.}
    \item \texttt{\textbf{SATISFACTORY:} The STATEMENT adequately meets this criterion but has noticeable areas for enhancement.}
    \item \texttt{\textbf{NEEDS IMPROVEMENT:} The STATEMENT partially meets this criterion but requires significant revisions.}
    \item \texttt{\textbf{UNSATISFACTORY:} The STATEMENT fails to meet this criterion.}
\end{itemize}

\texttt{
\newline
Provide the assessments (EXCELLENT/GOOD/SATISFACTORY/NEEDS IMPROVEMENT/UNSATISFACTORY) for each criterion without additional justification. Respond with the following structure:}
\begin{itemize}
    \item \texttt{\textbf{Factual Accuracy:} [Assessment]}
    \item \texttt{\textbf{Logical Coherence:} [Assessment]}
    \item \texttt{\textbf{Clarity:} [Assessment]}
    \item \texttt{\textbf{Relevance:} [Assessment]}
    \item \texttt{\textbf{Depth of Argumentation:} [Assessment]}
\end{itemize}

\\
\midrule[0.95pt]
%\rowcolor{beige}
\scriptsize
\texttt{\textbf{User}: Here is the question:}
\texttt{\textbf{[QUESTION]}
\newline}

\texttt{Here is the statement you need to evaluate:}
\texttt{\textbf{[STATEMENT]}}

\\
                    
\bottomrule[0.95pt]
\end{tabular}
\end{center}
\end{table}

\subsection{Consultant Component}
In cases where the model \(\mathcal{M}_{\text{SFT}}\) behaves consistently incorrectly for the input prompt \(x_i\), we employ the model \(\mathcal{M}_{\text{Base}}\) in a consultant role. Specifically, we provide the correct answer \(y_{i}\) to the LLM and request an explanation $\varepsilon_{i}$ supporting this answer.

\[
\mathcal{M}_{\text{Debater}}(x_i, y_i) \rightarrow \varepsilon_{i}
\]

\subsection{Consultant prompt}
\label{appendix: Consultant prompt}

\begin{table}[H]
\caption{\textbf{Consultant prompt}.}
\label{sample-table}
\begin{center}
\begin{tabular}{p{13cm}} % Adjust the width as needed
\toprule[0.95pt]
\textbf{Consultant prompt}\\
\midrule[0.95pt]
\midrule[0.95pt]
%\rowcolor{lightgray} 
\scriptsize

\texttt{\textbf{System}: You will act as an expert consultant. You will be shown a QUESTION and one OPTION to argue for. Your goal is to construct convincing arguments supporting your option as correct.
\newline \newline
Maintain a scholarly yet assertive tone, focusing on evidence-based analysis while crafting your arguments and answering questions. Avoid phrases like "In summary", "In conclusion", or similar at the end of your argument. Be as concise as possible. 
\newline \newline
Your argument must fulfill the following criteria:}
\begin{itemize}
    \item \texttt{\textbf{Truthfulness}: Any factual information mentioned in your argument must be true and accurate.}
    \item \texttt{\textbf{Logical Consistency}: Your argument must be logically sound, free of logical fallacies or contradictions.}
    \item \texttt{\textbf{Clarity}: Your argument directly addresses and answers the question. It is clear, concrete, concise, and well-structured.}
\end{itemize}

\\
\midrule[0.95pt]
%\rowcolor{beige}
\scriptsize
\texttt{\textbf{User}: Here is the question:}
\texttt{\textbf{[QUESTION]}
\newline}

\texttt{Here is option you need to back up:}
\texttt{\textbf{[OPTION]}}

\\
                    
\bottomrule[0.95pt]
\end{tabular}
\end{center}
\end{table}

\section{Prompts inference per task}
\label{appendix: Prompts inference per task}

\begin{table}[H]
\caption{\textbf{Prompt for LogiQA task}.}
\label{sample-table}
\begin{center}
\begin{tabular}{p{13cm}} % Adjust the width as needed
\toprule[0.95pt]
\textbf{Prompt for LogiQA task}\\
\midrule[0.95pt]
\midrule[0.95pt]
%\rowcolor{lightgray} 
\scriptsize

\texttt{\textbf{System}: You will be presented with a CONTEXT passage and a corresponding QUESTION with four answer CHOICES. 
Carefully read the passage to understand its content. Then, read the QUESTION and CHOICES thoroughly. Choose the correct CHOICE and explain your reasoning.
\newline \newline
Your response will consist of two parts: an EXPLANATION followed by your selected CHOICE.
\newline \newline
Enclose your explanation within tags as follows: \newline
<explanation>[Your EXPLANATION here]</explanation>
\newline \newline
Enclose your chosen choice (e.g., if the question has only 4 choices, then A, B, C, or D) within tags as follows: \newline
<choice>[Your CHOICE here]</choice>}

\\
\midrule[0.95pt]
%\rowcolor{beige}
\scriptsize
\texttt{\textbf{User}: 
Context:}
\texttt{\textbf{[CONTEXT]}
\newline}

\texttt{Question:}
\texttt{\textbf{[QUESTION]}\newline}

\texttt{Choices:}
\texttt{\textbf{[CHOICES]}\newline}
\\
                    
\bottomrule[0.95pt]
\end{tabular}
\end{center}
\end{table}

\subsection{Prompt for AQuA-Rat task}
\label{appendix: CoT prompt for LogiQA task}

\begin{table}[H]
\caption{\textbf{Prompt for AQuA-Rat task}.}
\label{sample-table}
\begin{center}
\begin{tabular}{p{13cm}} % Adjust the width as needed
\toprule[0.95pt]
\textbf{Prompt for AQuA-Rat task}\\
\midrule[0.95pt]
\midrule[0.95pt]
%\rowcolor{lightgray} 
\scriptsize

\texttt{\textbf{System}: You will be given a QUESTION along with multiple answer CHOICES, involving a math problem that requires step-by-step reasoning to determine the correct answer.
Carefully read the QUESTION and CHOICES. Choose the correct CHOICE and explain your reasoning.
\newline \newline
Your response will consist of two parts: an EXPLANATION followed by your selected CHOICE.
\newline \newline
Enclose your explanation within tags as follows: \newline
<explanation>[Your EXPLANATION here]</explanation>
\newline \newline
Enclose your chosen choice (e.g., if the question has only 4 choices, then A, B, C, or D) within tags as follows: \newline
<choice>[Your CHOICE here]</choice>}

\\
\midrule[0.95pt]
%\rowcolor{beige}
\scriptsize
\texttt{\textbf{User}: 
Context:}
\texttt{\textbf{[CONTEXT]}
\newline}

\texttt{Question:}
\texttt{\textbf{[QUESTION]}\newline}

\texttt{Choices:}
\texttt{\textbf{[CHOICES]}\newline}
\\
                    
\bottomrule[0.95pt]
\end{tabular}
\end{center}
\end{table}

\subsection{Prompt for ARC-Challenge task}
\label{appendix: CoT prompt for LogiQA task}

\begin{table}[H]
\caption{\textbf{Prompt for ARC-Challenge task}.}
\label{sample-table}
\begin{center}
\begin{tabular}{p{13cm}} % Adjust the width as needed
\toprule[0.95pt]
\textbf{Prompt for ARC-Challenge task}\\
\midrule[0.95pt]
\midrule[0.95pt]
%\rowcolor{lightgray} 
\scriptsize

\texttt{\textbf{System}: You will be presented a QUESTION with multiple answer CHOICES.
Carefully read the QUESTION and CHOICES. Choose the correct CHOICE and explain your reasoning.
\newline \newline
Your response will consist of two parts: an EXPLANATION followed by your selected CHOICE.
\newline \newline
Enclose your explanation within tags as follows: \newline
<explanation>[Your EXPLANATION here]</explanation>
\newline \newline
Enclose your chosen choice (e.g., if the question has only 4 choices, then A, B, C, or D) within tags as follows: \newline
<choice>[Your CHOICE here]</choice>}

\\
\midrule[0.95pt]
%\rowcolor{beige}
\scriptsize
\texttt{\textbf{User}: 
Context:}
\texttt{\textbf{[CONTEXT]}
\newline}

\texttt{Question:}
\texttt{\textbf{[QUESTION]}\newline}

\texttt{Choices:}
\texttt{\textbf{[CHOICES]}\newline}
\\
                    
\bottomrule[0.95pt]
\end{tabular}
\end{center}
\end{table}

\subsection{Prompt for OpenbookQA task}
\label{appendix: CoT prompt for LogiQA task}

\begin{table}[H]
\caption{\textbf{Prompt for OpenbookQA task}.}
\label{sample-table}
\begin{center}
\begin{tabular}{p{13cm}} % Adjust the width as needed
\toprule[0.95pt]
\textbf{Prompt for OpenbookQA task}\\
\midrule[0.95pt]
\midrule[0.95pt]
%\rowcolor{lightgray} 
\scriptsize

\texttt{\textbf{System}: You will be presented a QUESTION with multiple answer CHOICES.
Carefully read the QUESTION and CHOICES. Choose the correct CHOICE and explain your reasoning.
\newline \newline
Your response will consist of two parts: an EXPLANATION followed by your selected CHOICE.
\newline \newline
Enclose your explanation within tags as follows: \newline
<explanation>[Your EXPLANATION here]</explanation>
\newline \newline
Enclose your chosen choice (e.g., if the question has only 4 choices, then A, B, C, or D) within tags as follows: \newline
<choice>[Your CHOICE here]</choice>}

\\
\midrule[0.95pt]
%\rowcolor{beige}
\scriptsize
\texttt{\textbf{User}: 
Context:}
\texttt{\textbf{[CONTEXT]}
\newline}

\texttt{Question:}
\texttt{\textbf{[QUESTION]}\newline}

\texttt{Choices:}
\texttt{\textbf{[CHOICES]}\newline}
\\
                    
\bottomrule[0.95pt]
\end{tabular}
\end{center}
\end{table}

\section{Generated Instructions}

\begin{table}[h]
\caption{\textbf{Distribution of anchor categories:} This table presents the distribution of the categories—Consistently Correct (CC), Consistently Incorrect (CI), and Variable (V)—across datasets used during the DPO alignment phase of 
 \( \mathcal{M}_{\text{Anchor}} \) .
}
\label{tab:anchor_distribution}
\begin{center}

\begin{tabular}{lccc}
\toprule
\textbf{Dataset} & \textbf{Category} & \textbf{ Samples} & \textbf{Ratio (\%)} \\
\midrule
\midrule

\multirow{3}{*}{\texttt{AQuA-Rat}} 
    & V & $1196$ & $41.17$ \\
    & CC & $1010$ & $34.77$ \\
    & CI & $699$ & $24.06$ \\
\midrule
\multirow{3}{*}{\texttt{ARC-Challenge}}
    & V & $62$ & $8.09$ \\
    & CC & $645$ & $84.20$ \\
    & CI & $59$ & $7.70$ \\
\midrule

\multirow{3}{*}{\texttt{LogiQA} }
    & V & $1251$ & $26.86$ \\
    & CC & $2487$ & $53.39$ \\
    & CI & $920$ & $19.75$ \\
\midrule
\multirow{3}{*}{\texttt{OpenbookQA}} 
    & V & $176$ & $5.13$ \\
    & CC & $3178$ & $92.60$ \\
    & CI & $78$ & $2.27$ \\
\bottomrule
\end{tabular}
\end{center}
\end{table}

\end{document}